%% file: main.tex
\documentclass{article}

\usepackage[final]{neurips_2022}

\usepackage[utf8]{inputenc} 
\usepackage[T1]{fontenc}    
\usepackage{hyperref}       
\usepackage{url}            
\usepackage{booktabs}       
\usepackage{amsfonts}       
\usepackage{nicefrac}       
\usepackage{microtype}      
\usepackage{xcolor}         

\usepackage{subcaption}
\usepackage{multirow, makecell}
\usepackage{amsmath}
\usepackage{amssymb}
\usepackage{mathtools}
\usepackage{amsthm}
\usepackage{bm}
\usepackage[capitalise]{cleveref}

\theoremstyle{plain}
\newtheorem{theorem}{Theorem}[section]

\theoremstyle{definition}

\theoremstyle{remark}

\newcommand{\thetab}{{\bm{\theta}}}

\newcommand{\phib}{{\bm{\phi}}}
\newcommand{\Sigmab}{{\mathbf{\Sigma}}}
\newcommand{\mub}{{\bm{\mu}}}
\newcommand{\deltab}{{\bm{\delta}}}
\newcommand{\epsilonb}{{\bm{\epsilon}}}
\newcommand{\zb}{{\mathbf{z}}}

\newcommand{\yobs}[1]{{y_{\text{obs}}^{#1}}}
\newcommand{\yobsb}{{\mathbf{y}_{\text{obs}}}}
\newcommand{\xobs}[1]{{x_{\text{obs}}^{#1}}}
\newcommand{\xobsb}{{\mathbf{x}_{\text{obs}}}}
\newcommand{\hatY}[1]{{\hat{Y}_{#1}}}
\newcommand{\hatYb}{{\mathbf{\hat{Y}}}}
\newcommand{\xsample}[1]{{x_{\text{sample}}^{#1}}}
\newcommand{\xsampleb}{{\mathbf{x}_{\text{sample}}}}

\renewcommand{\mid}{\,|\,}

\def\1{\bm{1}}
\DeclareMathOperator*{\argmax}{arg\,max}
\DeclareMathOperator*{\argmin}{arg\,min}

\title{Generative Posterior Networks for Approximately Bayesian Epistemic Uncertainty Estimation}

\author{%
\begin{tabular}[t]{ccc}
\bf Melrose Roderick                & \hfil &           \bf Felix Berkenkamp                        \\
Carnegie Mellon University          & \hfil &           Bosch Center for Artificial Intelligence    \\
\texttt{mroderick@cmu.edu}          & \hfil &           \texttt{Felix.Berkenkamp@de.bosch.com}      \\
\\ \\
\bf Fatemeh Sheikholeslami          & \hfil &           \bf Zico Kolter                                 \\
Amazon Alexa AI                     & \hfil &           Carnegie Mellon University                  \\
\texttt{f.sheikholeslami@gmail.com} & \hfil &           \texttt{zkolter@cs.cmu.edu}                 \\
\end{tabular}
}

\begin{document}

\maketitle

\begin{abstract}
In many real-world problems, there is a limited set of training data, but an abundance of unlabeled data.
We propose a new method, Generative Posterior Networks (GPNs), that uses unlabeled data to estimate epistemic uncertainty in high-dimensional problems.
A GPN is a generative model that, given a prior distribution over functions, approximates the posterior distribution directly by regularizing the network towards samples from the prior.
We prove theoretically that our method indeed approximates the Bayesian posterior and show empirically that it improves epistemic uncertainty estimation and scalability over competing methods.
\end{abstract}

\section{Introduction}

In supervised learning tasks, the distribution of labeled training data often does not match exactly with the distribution of data the model will see at deployment.
This distributional shift can cause significant problems in safety-critical environments where mistakes can be catastrophic.
Ideally, we want our learned models to be able to estimate their epistemic uncertainty -- uncertainty deriving from lack of data samples.
Unfortunately, deep learning models struggle to estimate this type of uncertainty.
While there are many proposed methods for addressing this problem, epistemic uncertainty estimation in deep learning remains an open problem due to out of distribution (OOD) performance and scalability.

\begin{figure}[b]
\begin{center}
\begin{subfigure}{0.4\textwidth}
    \includegraphics[width=\textwidth]{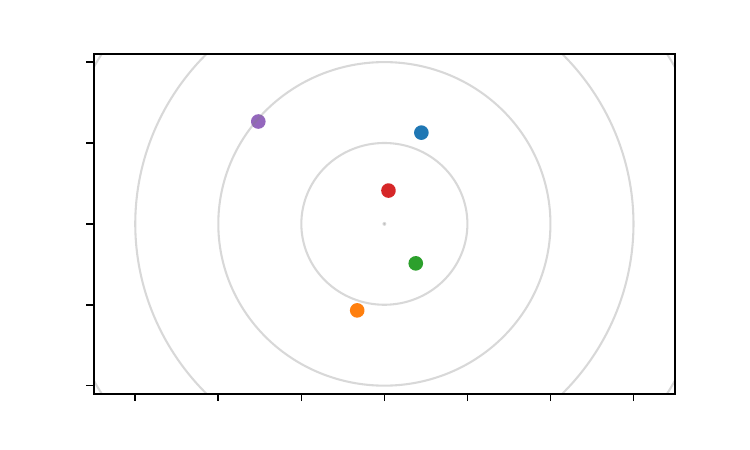}
    \caption{ 2D embedding samples }
\label{fig:example:a}
\end{subfigure}
\begin{subfigure}{0.4\textwidth}
    \includegraphics[width=\textwidth]{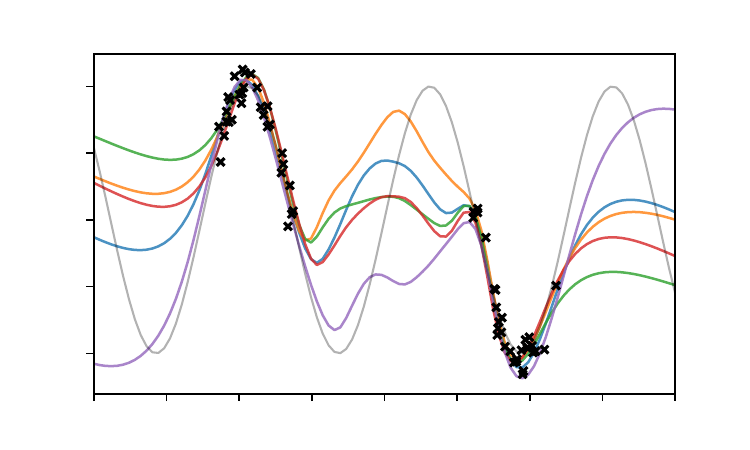}
    \caption{ Corresponding posterior samples }
\label{fig:example:b}
\end{subfigure}
\end{center}
\caption{Samples from a GPN using a 2D embedding trained on a simple sine-function.
On the top are samples from the embedding with corresponding posterior samples underneath.
Black `x's represent observed data points.}
\label{fig:example}
\end{figure}

Concurrently, many recent works have shown incredible performance using unlabeled data \citep{radford2019language, chen2020simple}.
While labeled training data is expensive to collect, in many real-world problems, such as image classification, there is an abundance of unlabeled data.
Moreover, this unlabeled data is often much more representative of the deployment distribution.
In this work, we propose a new method for estimating epistemic uncertainty that leverages this unlabeled data.

In many real-world tasks, labeled data is limited and expensive to obtain, while unlabeled data is plentiful.
Moreover, distribution of labeled data often does not match the desired test distribution.
This challenge has led to a growing literature in Domain Adaptation (DA) and Test-Time Adaptation (TTA).
Both of these problem settings assume access to a set of unlabeled data from the test distribution that is used to improve test performance.
However, if the test distribution is too far from the training distribution, DA and TTA approaches will still have high errors on the test distribution.
For such cases, understanding the model's epistemic uncertainty is vital.

Our work builds off of the extensive ensembling literature.
Neural network ensembling is a method that predicts epistemic uncertainty by looking at the disagreement between the ensemble members.
Additionally, under the right regularization, ensembles approximate samples from the Bayesian posterior \citet{pearce2020uncertainty, he2020bayesian}.
However, these methods have one main drawback, namely that each new sample from the posterior requires training a new network from scratch.

To address this challenge, we introduce Generative Posterior Networks (GPNs), a generative neural network model that directly approximates the posterior distribution by regularizing the output of the network towards samples from the prior distribution.
By learning a low-dimensional latent representation of the posterior, our method can quickly sample from the posterior and construct confidence intervals (CIs).
We prove that our method approximates the Bayesian posterior over functions and show empirically that our method can improve uncertainty estimation over competing methods.

\section{Related Work}

There are two categories of uncertainty in the context of probabilistic modeling: aleatoric uncertainty and epistemic uncertainty.
Aleatoric uncertainty refers to uncertainty inherent in a system, even when the true parameters of the system are known, while epistemic uncertainty refers to modeling uncertainty that can be reduced through collecting more data \citep{der2009aleatory, gal2016uncertainty}.
Deep learning models have shown impressive performance in quantifying aleatoric uncertainty when given enough data.
On the other hand, despite many methods seeking to estimate epistemic uncertainty due to its well-motivated applications in Bayesian Optimization \citep{snoek2012practical,springenberg2016bayesian}, Reinforcement Learning 
\citep{curi2020OptimisticModel}, and adversarial robustness \citep{stutz2020confidence}, quantifying it remains a challenging problem for deep learning models.

GPs remain one of the most effective models for epistemic uncertainty estimation in low-dimensional, low-data settings.
While scalability can be improved with interpolation methods \citep{hensman2013gaussian, titsias2009variational, wilson2015kernel} and kernel learning methods \citep{wilson2016deep}, GPs tend to perform worse than deep learning models on high-dimensional problems.
Spectral-normalized Neural Gaussian Processes (SNGP) \citep{liu2020simple} offers a potential solution to these issues.
By performing spectral-normalization on the deep kernel, \cite{liu2020simple} show that the learned kernel is distance preserving.
While SNGP shows impressive out of distribution prediction performance, we show in our experiments that the learned posterior does not adequately capture the true posterior.

\citet{van2020uncertainty} offer a unique approach to epistemic uncertainty estimation in the classification setting called deterministic uncertainty quantification (DUQ).
By learning a high-dimensional centroid for each class, DUQ estimates uncertainty by measuring the distance to the nearest centroid.
While this works well in classification settings, there is not a clear extension to regression settings.

\looseness=-1
Another way to formulate epistemic uncertainty estimation is as a Bayesian Inference problem.
Specifically, given a prior distribution over functions and some labeled data, the goal is to construct a posterior distribution in the Bayesian sense.
Because this posterior distribution is usually intractable to compute exactly, approximate sampling methods are used instead, such as Markov-Chain Monte Carlo (MCMC) \citep{tanner1987calculation} and Variational Inference \citep{blei2017variational}.
These methods, however, tend to struggle in high dimensional problems due to computational complexity and restrictive assumptions.
%
%
Bayesian Neural Networks (BNNs) \citep{blundell2015weight} and Bayesian Dropout \citep{blundell2015weight} are alternative methods for approximating Bayesian Inference on neural networks.
On high-dimensional problems, however, these methods tend to over-estimate uncertainty.

Neural network ensembling is another commonly used epistemic uncertainty estimation technique, which is closely related to our proposed method in this work.
This technique involves training a number of completely separate neural networks on the same data to form an ensemble.
Because of the stochasticity of initialization and the training process, each member of the ensemble should be a slightly different function.
Specifically, the ensemble members should produce similar outputs in data-dense regions of the state-space and dissimilar outputs in data-sparse regions.
\citet{pearce2020uncertainty} prove that, with a specific regularization, each ensemble member becomes an approximate sample from the function posterior.
\citet{he2020bayesian} proved a similar result using neural tangent kernels.
Ensembling methods are particularly attractive to practitioners because they require very little fine-tuning and work surprisingly well in practice.
However, these methods have one main drawback, namely that each new sample from the posterior requires training a new network from scratch.
Our method, on the other hand, seeks to construct a generative posterior model using similar regularization techniques to \citet{pearce2020uncertainty}, allowing for quick sampling from the posterior.





\begin{figure*}[t]
\begin{center}
\includegraphics[width=0.9\textwidth, trim=100 0 100 0, clip]{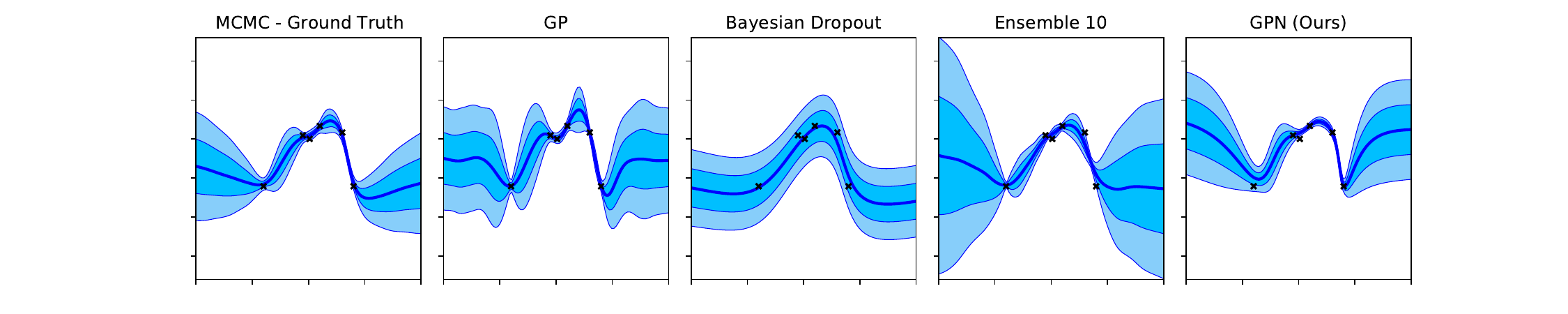}
\end{center}
\caption{Predicted posterior distributions of different methods using the same observed data.
}
\label{fig:small_scale}
\end{figure*}

\section{Problem Statement and Background}

We consider the problem of Bayesian Inference on the parameters of $\thetab$ of some function $f(\cdot; \thetab)$.
We assume we have some known prior over parameters $P_{\text{prior}} (\thetab) \sim \mathcal{N}(\mu_{\text{prior}}, \Sigmab_{\text{prior}})$ as well as access to a noisy dataset $(\xobsb, \yobsb) = \{(\xobs{1}, \yobs{1}) \ldots, (\xobs{N}, \yobs{N})\}$, where the observations $\yobs{i} = f(\xobs{i}; \thetab) + \epsilon$ depend on an unknown parameter $\thetab$ and are corrupted by Gaussian noise $\epsilon \sim \mathcal{N}(0, \sigma_{\epsilon})$.
We can then define the data-likelihood as
\begin{equation}
    P_{\text{like}}(\yobsb \mid \thetab, \xobsb) = \prod_i \mathcal{N}(\yobs{i} \mid f(\xobs{i}; \thetab), \sigma_{\epsilon}) .
\end{equation}
The goal is to approximate samples from the posterior,
\begin{equation}
    P_{\text{post}}(\thetab \mid \yobsb, \xobsb) \propto P_{\text{like}}(\yobsb \mid \thetab, \xobsb) P_{\text{prior}} (\thetab).
\end{equation}
For ease of notation, we drop the $\xobsb$ from the likelihood.

For our analysis, it is convenient if the posterior distribution is Gaussian.
To show this, we exploit that the product of two Gaussian PDFs is proportional to a Gaussian PDF.
We make a distinction here between the data-likelihood, a function of the observed data $\yobsb$, and the parameter-likelihood, a function of the parameters $\thetab$, defined as follows:
\begin{equation}
    P_{\text{param-like}}(\thetab ; \yobsb) = P_{\text{like}}(\yobsb \mid \thetab) .
\end{equation}
The parameter-likelihood PDF is equivalent to the data-likelihood PDF for all values of $\yobsb$ and $\thetab$, but is notably a function of the parameters $\thetab$.
For a more detailed explanation of the distinction between the parameter and data-likelihoods, see Definition 1 from \citet{pearce2020uncertainty}.

When the function $f$ is linear, it can be shown that the parameter-likelihood is also Gaussian,
\begin{equation}
    P_{\text{param-like}}(\thetab ; \yobsb) \propto \mathcal{N}(\mub_{\text{like}}, \Sigmab_{\text{like}}).
\end{equation}
While this is not true in general, \citet{pearce2020uncertainty} assume that the parameter likelihood is Gaussian so that the posterior also becomes Gaussian,
\begin{gather}
\begin{aligned}
    P_{\text{post}}(\thetab \mid \yobsb)
    &\propto P_{\text{param-like}}(\thetab ; \yobsb) P_{\text{prior}} (\thetab) \\
    &\propto \mathcal{N}(\mub_{\text{like}}, \Sigmab_{\text{like}}) \mathcal{N}(\mub_{\text{prior}}, \Sigmab_{\text{prior}})
    = \mathcal{N}(\mub_{\text{post}}, \Sigmab_{\text{post}}),
\end{aligned}
\end{gather}
where the posterior covariance and mean are given by
\begin{align}
    \mub_{\text{post}} &= \Sigmab_{\text{post}} \Sigmab_{\text{like}}^{-1} \mub_{\text{like}} + \Sigmab_{\text{post}} \Sigmab_{\text{prior}}^{-1} \mub_{\text{prior}} , \quad \quad
    \Sigmab_{\text{post}} = \left ( \Sigmab_{\text{like}}^{-1} + \Sigmab_{\text{prior}}^{-1} \right)^{-1} .
\end{align}
Note, however, that after the reformulation of the problem in Section~\ref{sec:theory}, we will show the parameter likelihood is always Gaussian, so we no longer require this assumption for our theoretical analysis.


\subsection{Randomized MAP sampling (RMS)}

Since the values $\mub_{\text{like}}$ and $\Sigmab_{\text{like}}$ are intractable to compute in practice, the challenge of computing the posterior distribution remains.
However, \citet{pearce2020uncertainty} showed how we can use a method called Randomized MAP Sampling (RMS) to approximately sample from the posterior distribution.
The idea behind RMS is to use the fact the parameters that minimize the regularized MSE loss are equal to the maximum a posteriori (MAP) solution for the posterior.
Thus, using optimization techniques, like gradient descent, we can easily find MAP solutions.
However, the challenge of sampling from the posterior remains.
Instead, RMS randomly samples shifted prior \textit{distributions} such that the resulting MAP solutions form samples from the original posterior.


Specifically, RMS samples ``anchor points'' $\thetab_{\text{anc}}$ from some distribution $\thetab_{\text{anc}} \sim \mathcal{N}(\mub_{\text{anc}}, \Sigmab_{\text{anc}})$.
Each anchor point corresponds to the mean of a shifted prior distribution $P_{\text{anc}}(\thetab) = \mathcal{N}(\thetab_{\text{anc}}, \Sigmab_{\text{anc}})$.
Now we define the MAP function which finds the MAP solution using this shifted prior:
\begin{equation}
    \theta_{\text{MAP}} (\thetab_{\text{anc}}) = \argmax_{\thetab} P_{\text{like}}(\yobsb \mid \thetab) P_{\text{anc}} (\thetab) .
\end{equation}
By sampling different anchor points, we can then construct a probability distribution over MAP solutions $P(\theta_{\text{MAP}} (\thetab_{\text{anc}}))$, which can be shown to be Gaussian.
Using moment matching, \citet[Theorem 1]{pearce2020uncertainty} shows that if we choose $\mub_{\text{anc}} = \mub_{\text{prior}}$ and $\Sigmab_{\text{anc}} = \Sigmab_{\text{prior}} + \Sigmab_{\text{prior}} \Sigmab_{\text{like}}^{-1} \Sigmab_{\text{prior}}$, then $P(\theta_{\text{MAP}} (\thetab_{\text{anc}})) = P_{\text{post}}(\thetab \mid \yobsb)$.
Using the approximation $\Sigmab_{\text{anc}} \approx \Sigmab_{\text{prior}}$, a sample from posterior, $\thetab_{\text{post}}$, can be collected by sampling $\thetab_{\text{anc}} \sim \mathcal{N}(\mub_{\text{anc}}, \Sigmab_{\text{anc}})$ and optimizing:
\begin{equation}
    \thetab_{\text{post}} = \argmax_{\thetab} \log P_{\text{param-like}}(\thetab ; \yobsb) + \log P_{\text{anc}} (\thetab) .
\end{equation}

\section{Generative Posterior Networks}

While RMS can be used to sample from the posterior, every new sample requires solving a new optimization problem.
Instead, we propose to learn the MAP function itself.

Let $g$ be a neural network parameterized by some vector, $\phi$, that takes as input a sample from the anchor distribution, $\thetab_{\text{anc}} \sim P_{\text{anc}}(\thetab_{\text{anc}})$, and outputs $\theta_{\text{MAP}} (\thetab_{\text{anc}})$.
If we assume that $g$ has enough expressive power to represent the MAP function, then $g(\thetab_{\text{anc}}; \phi) = \theta_{\text{MAP}} (\thetab_{\text{anc}})$ if:
\begin{equation}
\begin{aligned}
    \phi = \argmax_{\phi} \mathbb{E}_{\thetab_{\text{anc}}} 
    \big[
    \hphantom{+}\log P_{\text{like}}(\yobsb \mid g(\thetab_{\text{anc}}; \phi)) + \log P_{\text{anc}} (g(\thetab_{\text{anc}}; \phi)) \big] .
\end{aligned}
\end{equation}

In practice, however, this is a very hard optimization problem for two main reasons:
(1) Optimizing in parameter space tends to be challenging for gradient-descent optimization.
(2) The space of all parameters $\thetab_{\text{anc}}$ is not only very large, but there are also many parameters $\thetab_{\text{anc}}$ that map to the same outputs for all inputs.
Moreover, we would expect the posterior parameters to be highly correlated and, therefore, need less expressive power to be represented.

To address these two problems, we first change the problem definition slightly to find the posterior in \textit{output} space as opposed to \textit{parameter} space.
We prove in the next section that, by changing our loss function, we can still use RMS to approximate the posterior in output space.
Next, we show practically how we construct a low-dimensional representation of the anchor distribution.


\subsection{Theoretical Results}
\label{sec:theory}

As mentioned above, optimizing a generative model to output parameters can be challenging.
Instead, we reformulate the Bayesian Inference problem to consider the posterior in output space.
Concretely, consider some discretization of the input space, $\xsampleb = \{\xsample{1}, \ldots, \xsample{M} \}$.
Let $\hatY{i} = f(\xsample{i}; \thetab)$ be a set of sample points and $\hatYb = \{\hatY{1}, \ldots, \hatY{M}\}$ be a transformation of the random variable $\thetab$, where each element is the output of the function $f$ parameterized by $\thetab$ evaluated at points $\xsampleb$.
Because there is a many-to-one mapping between $\thetab$ and $\hatYb$, the prior and likelihood of $\hatYb$ can be expressed as an integral over all possible parameters $\thetab$:
\begin{align}
    P(\hatYb) &= \int_{\thetab} \1_{\hatY{i} = f(\xsample{i}; \thetab) \forall i} P_{\text{prior}} (\thetab) d\thetab , \\
    P(\yobsb \mid \hatYb) &\propto \int_{\thetab} \1_{\hatY{i} = f(\xsample{i}; \thetab) \forall i} P_{\text{like}} (\yobsb \mid \thetab) P_{\text{prior}}(\thetab) d\thetab \notag.
\end{align}
Note that samples from the prior of $\hatYb$ can be easily obtained by sampling $\thetab \sim P_{\text{like}} (\thetab)$ and computing $\hatY{i} = f(\xsample{i}; \thetab)$ for all $i$.
The goal in this reformulated problem is to find the posterior $P(\hatYb \mid \yobsb) \propto P(\yobsb \mid \hatYb) P (\hatYb)$.

As before, we estimate the posterior using RMS and define a new MAP function in the output space:
\begin{equation}
    \hat{Y}_{\text{MAP}} (\hatYb_{\text{anc}}) = \argmax_{\hatYb} P(\yobsb \mid \hatYb) P_{\text{anc}} (\hatYb_{\text{anc}})
\end{equation}
for some anchor distribution $\hatYb_{\text{anc}} \sim P_{\text{anc}}(\hatYb_{\text{anc}}) = \mathcal{N}(\mub_{\text{anc}}, \Sigmab_{\text{anc}})$.
We now need to show that, with the correct choice of $\mub_{\text{anc}}$ and $\Sigmab_{\text{anc}}$, $P(\hat{Y}_{\text{MAP}} (\hatYb_{\text{anc}})) = P(\hatYb \mid \yobsb)$.

To simplify the analysis, we assume that, for any sample inputs, $\xsampleb$, the output prior is approximately normally distributed $P(\hatYb) = \mathcal{N}(\mub_{\hatYb}, \Sigmab_{\hatYb})$.
Since the parameters are normally distributed, this approximation becomes exact when either $f$ is linear or the final layer is infinitely wide.

\begin{theorem}
\label{thm}
Let $f$ be some neural network parameterized by $\thetab \sim \mathcal{N}(\mub_{\text{prior}}, \Sigmab_{\text{prior}})$ such that, for any inputs $\mathbf{x}$, the outputs $\mathbf{y} = f(\mathbf{x}; \thetab)$ are jointly Gaussian.
Let $\xsampleb$ be some vector of inputs sampled from $\mathcal{X}^M$.
Define $\hatYb := [f(x_1; \thetab), \ldots, f(x_M; \thetab)]$ as a transformation of the random variable $\thetab$.
%
We assume access to some function which takes as input the center of a shifted distribution and outputs the MAP estimate of $\hatYb$, $\hat{Y}_{\text{MAP}}(\hatYb_{\text{anc}})$.

If we choose the distribution over $\hatYb_{\text{anc}}$ to be $P(\hatYb_{\text{anc}}) = \mathcal{N}(\mub_{\text{anc}}, \Sigmab_{\text{anc}})$, where $\mub_{\text{anc}} = \mub_\hatYb$ and $\Sigmab_{\text{anc}} = \Sigmab_\hatYb + \Sigmab_\hatYb \Sigmab_{\text{like}}^{-1} \Sigmab_\hatYb$, then $P(\hat{Y}_{\text{MAP}}(\hatYb_{\text{anc}})) = P(\hatYb \mid \yobsb)$.
\end{theorem}

\looseness=-1
The full proof is available in \cref{ap:proof_thm}, but we provide the high-level intuition here.
First, we need to show that the likelihood function $P(\yobsb \mid \hatYb)$ is approximately normally distributed with respect to the vector $\hatYb$.
Recall that each observed data point, $\yobs{i}$, is equal to a noisy output of $f$ for some parameter $\thetab$, $\yobs{i} = f(\xobs{i}; \thetab) + \epsilon$.
Using our assumption that the outputs of our function $f(\cdot; \thetab)$ are normally distributed, we can show that $\yobsb$, $\hatYb$, and $f(\xobs{i}; \thetab)$ are jointly Gaussian.
By taking a marginal and then a conditional of this joint Gaussian, we get that $P(\yobsb \mid \hatYb)$ is Gaussian distributed.

Now, since we have that both the prior and likelihood of $\hatYb$ can be expressed as Gaussian functions of $\hatYb$, we also have that the posterior is Gaussian.
As before, we just need to show that the distribution of MAP solutions $P(\hat{Y}_{\text{MAP}}(\hatYb_{\text{anc}}))$ is also Gaussian, then apply moment matching to find values for $\mub_{\text{anc}}$ and $\Sigmab_{\text{anc}}$ where $P(\hat{Y}_{\text{MAP}}(\hatYb_{\text{anc}}))) = P(\hatYb \mid \yobsb)$.

\subsection{Practical Implementation}
\label{sec:practical_implementation}
In practice, we follow \citet{pearce2020uncertainty} and instead use the approximation $\Sigmab_{\text{anc}} \approx \Sigmab_{\hatYb}$ to make it tractable to sample from the anchor distribution.
\citet{pearce2020uncertainty} argue that this approximation, in general, causes RMS to only over-estimate the posterior variance.

With this assumption, the anchor distribution $P_{\text{anc}}(\hatYb_{\text{anc}})$ becomes approximately equal to the prior distribution $P(\hatYb)$.
Recall that, to sample from the output prior, we simply sample a set of parameters $\thetab \sim P_{\text{prior}}(\thetab)$ and evaluate the function $f$ at all evaluation points $\xsampleb$.
Now, if we want to learn the MAP function in the output space, we can construct a neural network $g$, parameterized by some vector $\phi$, that takes a sample from the output anchor distribution, $\hatYb_{\text{anc}} \sim \mathcal{N}(\mub_{\text{anc}}, \Sigmab_{\text{anc}})$, along with the evaluation points $\xsampleb$, and outputs $\theta_{\text{MAP}} (\thetab_{\text{anc}})$.
If we assume that $g$ has enough expressive power to represent the MAP function, then $g(\hatYb_{\text{anc}}; \phib) = \hat{Y}_{\text{MAP}} (\hatYb_{\text{anc}})$, if $\phib$ is equal to:
\begin{gather}
\begin{aligned}
    \argmax_{\phib} \mathbb{E}_{\thetab_{\text{anc}}}
    \log P_{\thetab}(\yobsb \mid g(\xsampleb, \thetab_{\text{anc}}; \phib)) 
    + \log P_{\text{anc}} (g(\xsampleb, \thetab_{\text{anc}}; \phib))
\end{aligned}
\end{gather}
where $\thetab_{\text{anc}} \sim P_{\text{prior}}(\thetab_{\text{anc}})$.
This can be re-written as:
\begin{gather}
\begin{aligned}
    \argmin_{\phib} \mathbb{E}_{\thetab_{\text{anc}}} \sum_{i=1}^N
    \Big(& \|\yobs{i} - g(\xobs{i}, \thetab_{\text{anc}}; \phib)\|_2^2 
    + \sigma_{\epsilon}^2 \deltab^T \Sigmab_{\hatYb}^{-1} \deltab \Big )
\end{aligned}
\end{gather}
where $\delta_j = g(\xsample{j}, \thetab_{\text{anc}}; \phib) - f(\xsample{j}; \thetab_{\text{anc}})$.
See Appendix~\ref{ap:proof_loss} for a step-by-step derivation.

In practice, of course, it is expensive to compute the regularization term, $\frac{1}{N}\sigma_{\epsilon}^2 \deltab^T \Sigmab_{\hatYb}^{-1} \deltab$ for a large set of sample points $\xsampleb$.
Instead, we approximate this term using a $k$-size minibatch of sample points from the input space $\mathcal{X}$,
$\xsample{1}, \ldots, \xsample{k} \sim P_{\text{sample}}(x)$
where $P_{\text{sample}}(x)$ is a domain-specific distribution of interest, such as an unlabeled dataset.
Because this minibatch is small and the covariance between two samples grows with their L2-distance, the off-diagonal terms of $\Sigmab_{\hatYb}$ are near zero almost always for high-dimensional problems.
However, we found that, in practice, assuming that the samples of $\hatYb$ are independent did not significantly degrade performance.
Using this approximation, our optimization problem becomes:
\begin{equation}
\label{eq:output_reg}
    \argmin_{\phib} \mathbb{E}_{\thetab_{\text{anc}}} \sum_{i=1}^N \|\yobs{i} - g(\xobs{i}, \thetab_{\text{anc}}; \phib)\|_2^2 + \beta \| \delta \|_2^2
\end{equation}
where $\thetab_{\text{anc}} \sim P_{\text{prior} }(\thetab_{\text{anc}})$, $\xsample{} \sim \text{Unif}(\mathcal{X})$, $\delta{} = g(\xsample{}, \thetab_{\text{anc}}; \phib) - f(\xsample{}; \thetab_{\text{anc}})$, $\beta$ is a hyper-parameter.

In order for $g$ to have enough expressive power to represent the full MAP function, it may require the size of the parameters $\phib$ to be much larger than $\thetab$.
However, after applying the low-dimensional approximation of $\thetab_{\text{anc}}$ described below, in our experiments, we only used roughly double the number of parameters compared to each member of the ensemble method.
And since the ensemble we used has 10 members, the total number of parameters needed for our method is significantly smaller.

\subsection{Low-Dimensional Embedding}

As mentioned above, optimizing the generative function $g$ over the the space of all anchor parameters $\thetab_{\text{anc}}$ is impractical, as this space is too high dimensional.
Instead, we seek to learn a low-dimensional representation of the anchor parameters.
There are two key reasons why we would expect to be able to decrease the representational power of the anchor distribution and maintain the same of fidelity in the posterior estimation.
(1) Because of the nature of neural networks, there are many settings of parameters $\thetab$ that would result in the same output vector $\hatYb$.
Because we are focusing on the posterior of $\hatYb$, we need less representational power to model $\hatYb$.
(2) And, maybe more importantly, because the posterior parameters are highly correlated, we would expect to need less expressive power to represent the posterior distribution than the prior distribution.

Thus, we would like to construct a simpler estimate of the prior space using a low-dimensional embedding vector, $\zb$.
That is, we want our generative model to take as input the embedding vector $\zb \sim \mathcal{N}(0, I)$, instead of $\thetab_{\text{anc}}$, in order to estimate the MAP function.
To do this, we need to learn a mapping from anchor parameters $\thetab_{\text{anc}}$ to embedding vectors $\zb$.
For our experiments, we used a 1-1 embedding scheme where we sample $k$ parameters from the true prior $\thetab_1, \ldots, \thetab_k \sim P_{\text{prior}}(\thetab)$ and $k$ independent samples from our embedding $\zb_1, \ldots, \zb_k$.
We then jointly optimize over $\phib$ and $\zb_1, \ldots, \zb_k$ as follows:
\begin{align}
    \argmin_{\phib, \zb_1, \ldots, \zb_k} \mathbb{E}_{\xsampleb} \sum_{j=1}^k &\sum_{i=1}^N \|\yobs{i} - g(\xobs{i}, \zb_j + \epsilonb; \phib)\|_2^2 
    + \beta \| \delta \|_2^2 + \mathcal{L}_{\text{reg}} (\zb_1, \ldots, \zb_k),
\end{align}
where $\delta_j = g(\xsample{j}, \zb_j + \epsilonb; \phib) - f(\xsample{j}; \thetab_j)$, $\xsample{j} \sim P_{\text{sample}}(x)$, $\epsilonb \sim \mathcal{N}(0, I)$ is a noise injection vector that improves the smoothness of interpolations between embedding vectors, and $\mathcal{L}_{\text{reg}} (\zb_1, \ldots, \zb_k)$ is a regularizer to keep $\zb_1, \ldots, \zb_k$ roughly normally distributed, which allows us to easily sample from the embedding space.
For our experiments we use the KL divergence between $\zb$ and the normal distribution, $\mathcal{L}_{\text{reg}} (\zb_1, \ldots, \zb_k) = D_{\text{KL}} (\mathcal{N}(\bar{\zb}, s_{\zb}), \mathcal{N}(0, 1))$.
Figure~\ref{fig:example} illustrates how we can sample from this embedding space to construct posterior functions.

\subsection{Classification}

Our method uses the assumption that the prior over $\hatYb$ is approximately normally distributed.
Of course, this approximation is only exact when either the network is linear or the final layer is infinitely wide and does not have an activation function (assuming the prior parameters are normally distributed).
Thus, in classification tasks, where it is beneficial to a soft-max to the final output, this assumption is violated.
To get around this, we add the anchor loss to the pre-softmax outputs for classification tasks.

\begin{table*}[t]
\begin{center}
\caption{
Superconductor regression \citep{hamidieh2018data} results.
}
\begin{tabular}{ l c c c c c}
\toprule
Methods & In Dist. Loss & In Dist. CI-width & OOD CI-correct & OOD-detect. AUC \\
\midrule
Dropout & 0.11 & 0.32 & 15.9\% & 0.58 \\
DKL GP & 0.13 & 0.74 & 34.1\% & 0.50 \\
SNGP & 0.12 & 1.45 & 86.0\% & 0.82 \\
Ensemble - PR & 0.12 & 0.26 & 84.7\% & 0.92 \\ 
Ensemble - OR & 0.12 & 0.26 & 76.5\% & 0.95 \\
GPN (Ours) & 0.12 & 0.32 & 84.8\% & \textbf{0.96} \\
\bottomrule
\end{tabular}
\label{tab:reg_results}
\end{center}
\end{table*}


\begin{table*}[t]
\begin{center}
\caption{MNIST results with Out of Distribution (OOD) evaluation on Fashion MNIST.
}
\begin{tabular}{ l c c c c c}
\toprule
Methods & In Dist. Accuracy & In Dist. Entropy & OOD Entropy & OOD-detect. AUC \\
\midrule
Dropout & 98.9\% & 0.009 & 0.444 & 0.97 \\
DKL GP & 98.9\% & 0.091 & 1.223 & 0.99 \\
SNGP & 98.5\% & 2.300 & 2.300 & 0.99 \\
Ensemble - PR & 99.1\% & 0.056 & 0.862 & 0.97 \\
Ensemble - OR & 99.1\% & 0.070 & 2.075 & \textbf{1.00} \\
GPN (Ours) & 99.1\% & 0.280 & 2.032 & \textbf{1.00} \\
\bottomrule
\end{tabular}
\label{tab:mnist_results}
\end{center}
\end{table*}

\begin{table*}[t]
\begin{center}
\caption{CIFAR-10 results with Out of Distribution (OOD) evaluation on SVHN.
}
\begin{tabular}{ l c c c c c}
\toprule
Methods & In Dist. Accuracy & In Dist. Entropy & OOD Entropy & OOD-detect. AUC \\
\midrule
Dropout & 64.8\% & 0.103 & 0.233 & 0.71 \\
DKL GP & 79.3\% & 0.027 & 0.372 & 0.76 \\
SNGP & 78.7\% & 0.375 & 0.847 & 0.79 \\
Ensemble - PR & 76.4\% & 0.596 & 0.640 & 0.49 \\
Ensemble - OR & 76.3\% & 0.468 & 2.090 & 0.91 \\
GPN (Ours) & 75.8\%& 0.629& 2.246& \textbf{0.96} \\
\bottomrule
\end{tabular}
\label{tab:cifar_results}
\end{center}
\end{table*}

\section{Experiments}
\label{sec:experiments}


The goal of our experiments is to illustrate the ability of our method to accurately model epistemic uncertainty for out of distribution data while retaining high performance on in distribution data.
To do so, similar to related work \citep{liu2020simple, van2020uncertainty}, we train the models using one dataset, ``In Distribution'', and evaluate the model on a different dataset, ``Out of Distribution'' (OOD).
Table~\ref{tab:datasets} in \cref{ap:exp_details} provides quick reference for what datasets were used in each task.

\paragraph{Small Scale Regression}
Our first experiment is a small 1-dimensional regression problem where we can easily compute the ground truth.
We provide each method with the same 6 observations.
We compute 100 samples from the approximate posterior for each method.
For ground truth, we use the Metropolis-Hastings MCMC algorithm.

\paragraph{High-Dimensional Regression}
For a high-dimensional regression task, we used a Superconductivity prediction dataset \citep{hamidieh2018data}, the goal of which is to predict the critical temperature of different superconductive materials based on 81 features. 
For the In Distribution and OOD datasets, we split the full dataset based on the target values.
Specifically, the In Distribution and OOD datasets contained data with a target values in the intervals $[13.9, +\infty)$ and $[0, 13.9)$, which is roughly $58\%$ of the data in distribution and $42\%$ out of distribution.
The unlabeled dataset we provide our method and the Output-Regularized ensemble method is the full training set.
Regularization hyper-parameters were chosen by running each method with 5 different settings, then choosing the model with both a low validation loss and high CI-correct on a validation set.

\paragraph{Classification}
We run two different classification experiments: for one, we use the MNIST as our In Distribution dataset and Fashion MNIST as the OOD dataset, and for the other, we use the CIFAR-10 dataset as our In Distribution dataset and SVHN as the OOD dataset.
The unlabeled dataset we provide our method and the Output-Regularized ensemble method consists of unlabeled data 50\% from the In Distribution and 50\% from the OOD datasets.
As in the regression experiments, we ran each method with 5 different regularization parameters, then chose the model with both a high validation accuracy and OOD prediction performance on a validation set.

\subsection{Baselines}

We compare our method against 4 competing Bayesian methods:
Bayesian Dropout \citep{gal2016dropout}, GPs, and anchor-regularized neural-network ensembles \citep{pearce2020uncertainty} with 10 ensemble members.
For the high-dimensional problems, we need to use approximate GPs.
Specifically, we use a grid-interpolated GP with Deep Kernel Learning (DKL) \citep{wilson2016deep} implemented with the GPyTorch library \citep{gardner2018gpytorch} and the Spectral-normalized Neural Gaussian Process (SNGP) method \citep{liu2020simple}.
For the ensembles, since every member requires a unique set of parameters (and anchor points) the number of ensemble members was limited by the memory capacity of our GPUs.

As described in Equation~\ref{eq:output_reg}, our method requires a function to sample unlabeled data points.
In low-dimensional problems, this can simply be a uniform sample from the interval of interest, for example $[-2, 2]$ in Figure~\ref{fig:small_scale}.
However, in high-dimensional problems, this requires access to an additional dataset of unlabeled training data.
From a practical perspective, this is a reasonable setup as many real-world datasets have far more unlabeled data than labeled data.

This additional data obviously favors our method over the baseline methods as our method is the only method that is able to take advantage of this unlabeled data.
To address this, we added an additional ensemble regularized using the method in Equation~\ref{eq:output_reg}; in other words, we regularize the outputs of the ensemble members towards outputs of sampled prior networks.
We denote this new method as an ``Output-regularized (OR)'' Ensemble to distinguish it from the ``Parameter-regularized'' (PR) Ensemble.
We show that despite having the access to the same additional unlabeled data, our method outperforms and scales better than this OR Ensemble method.

Each model uses the same 4-layer (Superconductor), 5-layer (MNIST), or 7-layer (CIFAR) architecture, except the ensembles, which use slightly smaller networks for each ensemble member (to fit on a single GPU).
Full experimental details are in \cref{ap:exp_details}.

\subsection{Metrics}

Because we are interested in epistemic uncertainty estimation, we cannot use common aleatoric uncertainty evaluation metrics, such as Expected Calibration Error (ECE) \citep{guo2017calibration}.
Instead, we evaluate each methods ability to construct useful posterior estimates using the following metrics.

\paragraph{OOD Detection}
One desired property of an epistemic uncertainty estimators is in predicting when a data point is outside the training distribution and, thus, any prediction made on such a data point will likely be incorrect.
For this reason, OOD detection is a common metric used in epistemic uncertainty literature \citep{liu2020simple, van2020uncertainty}.
For an ideal posterior distribution, we expect high sample variance on OOD data and low sample variance from In Distribution data.
Thus, in our experiments, we use posterior sample variance to detect OOD data.
Specifically, for each model, we use posterior variance over 100 samples from each model as a score to construct ROC curves and measure the measure the area under the curve (AUC).

\paragraph{Confidence Intervals / Entropy}
Another desired property of an epistemic uncertainty estimator for a regression task is in constructing confidence intervals.
We would expect such confidence intervals (CIs) to contain the true class with high probability.
We also expect that, on in distribution examples, confidence intervals are narrow.
To construct these intervals, we take 100 samples from each model on every data point in the test data and compute the range of the middle $95\%$ of the samples.
\textit{CI-correct} refers to the percentage of true labels that fall in this interval.
Note that, unlike the other methods tested, ensembles are not a generative model.
Thus, when trying to sample from the ensemble method, we can only sample as many functions as there are members of the ensemble; for our experiments this number is 10.

For classification, CIs are not as informative.
Instead, we look at the posterior sample entropy.
For an ideal posterior distribution, we expect entropy to be low for data inside the training set (data the model should have confidence on) and high for data outside the training set (which we expect the model to be uncertain on).
To measure this sample entropy, we take 100 samples from each model on every data point and take the average of the post-softmax outputs.
We then average this sample entropy for both the In Distribution and OOD datasets.

\begin{figure}[t]
    \centering
    \begin{subfigure}[b]{0.45\columnwidth}
        \centering
        \includegraphics[width=0.8\columnwidth, trim=0 0 0 0, clip]{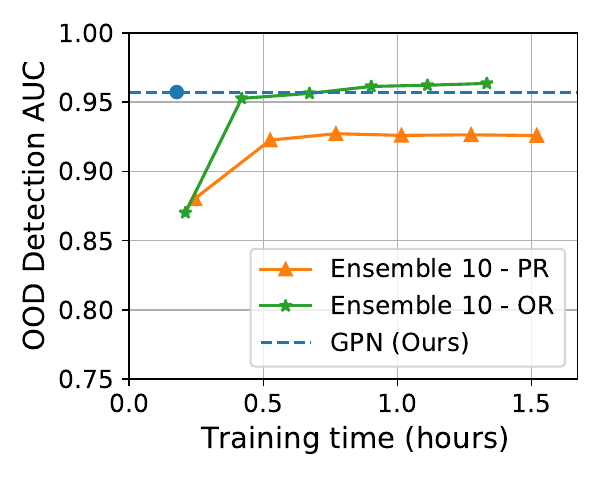}
        \caption{ Regression - Superconductor }
    \end{subfigure}
    \begin{subfigure}[b]{0.45\columnwidth}
        \centering
        \includegraphics[width=0.8\columnwidth, trim=0 0 0 0, clip]{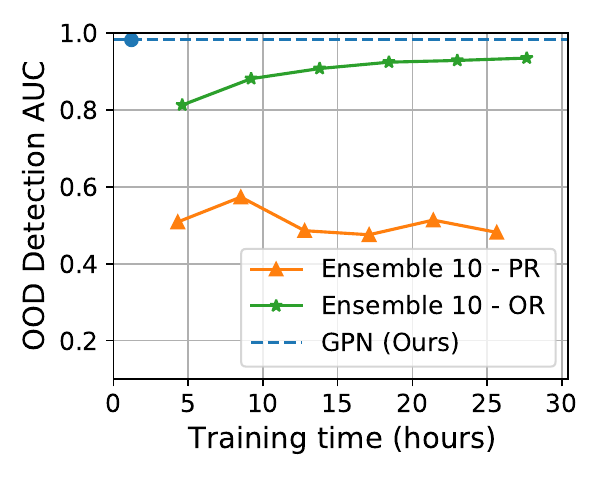}
        \caption{ Classification - CIFAR-10}
    \end{subfigure}
    \caption{OOD detection AUC vs. training time on the Superconductor and CIFAR-10 datasets for parameter and output-regularized ensembles and our method (GPN).}
    \label{fig:scalability}
\end{figure}

\paragraph{Scalability}
One key benefit of a generative model over an ensemble is efficiency: every sample from an ensemble method requires learning an entire new network from scratch.
Generative models, on the other hand, are able to produce new samples with a single pass through the network.
To illustrate the benefit of this efficiency in practice, we ran an experiment to measure OOD detection AUC vs. computation time for both the Superconductor and CIFAR-10 tasks.
%
We trained 2 (Superconductor) and 5 (CIFAR) ensemble members at a time and computed the OOD-detection performance of the cumulative combined ensembles (both parameter-regularized and output-regularized).
Each method is trained using an Nvidia 1080TI.
Figure~\ref{fig:scalability} shows our method is able to achieve high out of distribution prediction performance in significantly less computation time.

\subsection{Results}
\label{sec:results}
See Figure~\ref{fig:small_scale} and Tables~\ref{tab:reg_results}, \ref{tab:mnist_results}, and \ref{tab:cifar_results} for results.
ROC curves are presented in Figure~\ref{fig:roc} in \cref{ap:add_figs}.
While all methods are able to achieve high In Distribution performance (loss or accuracy), the GPN is able to consistently achieve the highest OOD-detection AUC.
The SNGP and OR-ensemble are also able to achieve high OOD-detection AUC, but have other drawbacks.
Specifically, SNGP has a very high CI-width on In Distribution data for the regression task and a very small contrast in sample entropy between the In Distribution and OOD data for the classification tasks.
And, while the OR-ensemble performs well, Figure~\ref{fig:scalability} shows GPN scales much better.
The PR-ensemble method performs surprisingly poorly in our experiments.
We think this is due to the challenge of finding a good prior distribution over weights.



Figure~\ref{fig:mnist} in \cref{ap:add_figs} shows estimated posteriors for each method on two informative test images, one from the In Distribution dataset and one from the OOD dataset, for both classification tasks.
While all methods predict the correct class with high precision on the In Distribution example, the GPN has a uniquely wide sample distribution on the OOD example, illustrating high predicted epistemic uncertainty.

\section{Conclusion}

In this paper we introduce Generative Posterior Networks (GPNs), a method to learn a generative model of the Bayesian posterior distribution by regularizing the outputs of the network towards samples of the prior.
We prove that under mild assumptions, GPNs approximate samples from the true posterior.
We then show empirically that our method significantly outperforms competing epistemic uncertainty predictions techniques on high-dimensional classification tasks and scales much better than ensembling methods, the closest performing baseline.

\section*{Acknowledgements}
Melrose Roderick was supported by a grant from the Bosch Center for AI.
This material is based upon work supported by the National Science Foundation Graduate Research Fellowship under Grant No.~DGE1745016.


\bibliography{references}
\bibliographystyle{icml2022}

\newpage
\appendix

\section{Additional Figures}
\label{ap:add_figs}

\begin{figure*}
\begin{center}
\includegraphics[width=0.95\textwidth, trim=0 0 0 0, clip]{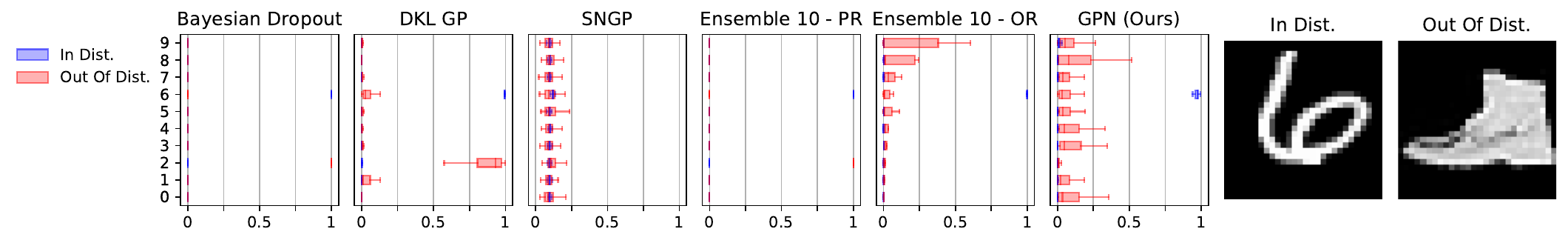}
\includegraphics[width=0.95\textwidth, trim=-46 0 0 0, clip]{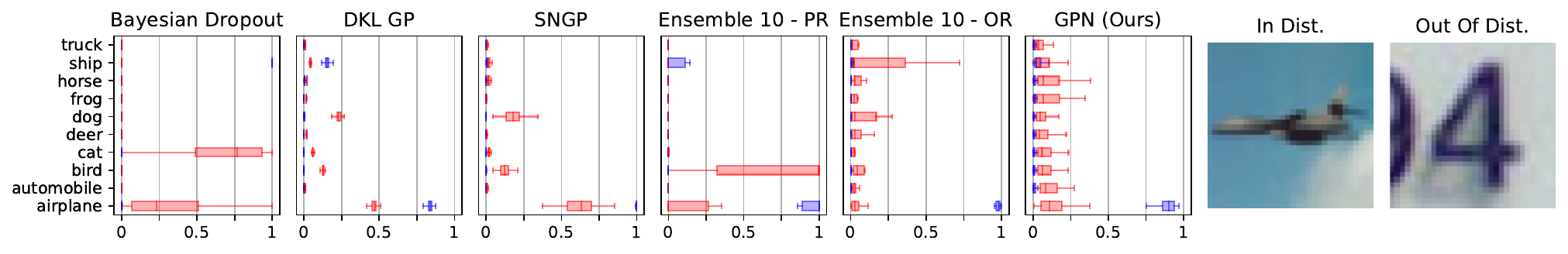}
\end{center}
\caption{Boxplots of 100 posterior samples from every method for 2 test images, one from the In Distribution dataset and one from the OOD dataset.
}
\label{fig:mnist}
\end{figure*}

Figure~\ref{fig:mnist} illustrates our methods ability to form credible posterior distributions.
Specifically, the figure shows two informative test images, one from the In Distribution dataset and one from the OOD dataset, for both MNIST/Fashion MNIST and CIFAR-10/SVHN, along with box plots of sampled PMFs from each of the learned models.
While all methods predict the correct label with high precision on the in distribution example, ours has a uniquely wide sample distribution on the OOD example, illustrating high predicted epistemic uncertainty.

\begin{figure}
    \centering
    \begin{subfigure}[b]{0.3\columnwidth}
        \includegraphics[width=\columnwidth, trim=0 0 0 0, clip]{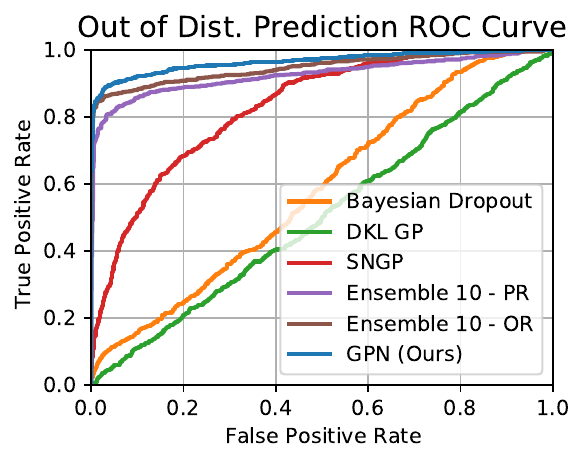}
        \caption{ Regression - Superconductor }
    \end{subfigure}
    \begin{subfigure}[b]{0.3\columnwidth}
        \includegraphics[width=\columnwidth, trim=0 0 0 0, clip]{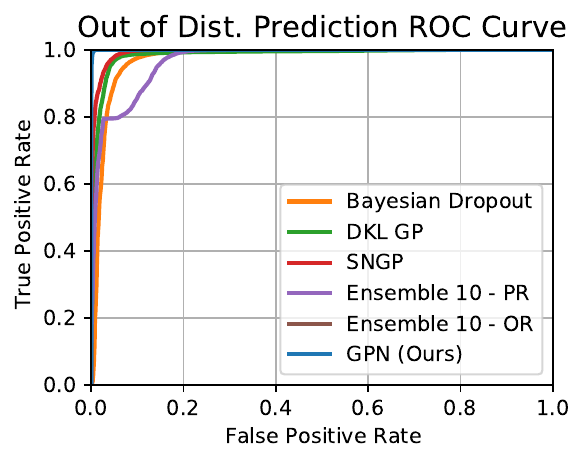}
        \caption{ Classification - MNIST }
    \end{subfigure}
    \begin{subfigure}[b]{0.3\columnwidth}
        \includegraphics[width=\columnwidth, trim=0 0 0 0, clip]{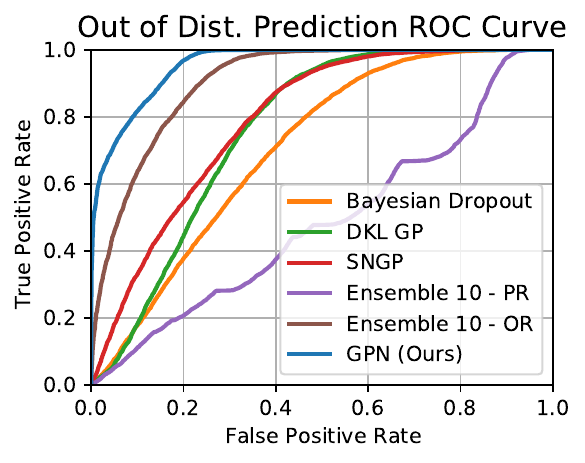}
        \caption{ Classification - CIFAR-10 }
    \end{subfigure}
    \caption{ROC curves for out of distribution prediction based on sample variance from 100 samples.}
    \label{fig:roc}
\end{figure}

Figures~\ref{fig:roc} show the full ROC curves for the OOD prediction tasks.

\section{Experiment Details}
\label{ap:exp_details}

\begin{table*}[t]
\begin{center}
\caption{
Training and test datasets used for each experiment task.
}
\begin{tabular}{ l c c c c }
\toprule
\multirow{2}{*}{Task} & \multicolumn{2}{c}{Training} & \multicolumn{2}{c}{Testing} \\
\cmidrule(r){2-3}
\cmidrule(r){4-5}
 & Labeled & Unlabeled & In Dist. & OOD \\
\midrule
Small Scale & 6 points & $\mathcal{U}[-2, 2]$ & N/A & $\mathcal{U}[-2, 2]$ \\
\midrule
Superconductor & \makecell{train set where \\ $y_i \in [13.9, +\infty)$} & full train set & \makecell{test set where \\ $y_i \in [13.9, +\infty)$} & \makecell{test set where \\ $y_i \in [0, 13.9)$} \\
\midrule
MNIST & MNIST train & \makecell{50\% MNIST train \\ 50\% F-MNIST train} & MNIST test & F-MNIST test \\
\midrule
CIFAR-10 & CIFAR-10 train & \makecell{50\% CIFAR-10 train \\ 50\% SVHN train} & CIFAR-10 test & SVHN test \\
\bottomrule
\end{tabular}
\label{tab:datasets}
\end{center}
\end{table*}

\paragraph{Experiment datasets}
Table~\ref{tab:datasets} details the labeled and unlabeled data provided to the agent at training time and the In Distribution and OOD data used for evaluation at test time.

\paragraph{Network Architectures}
For the superconductor regression problem, we used a 4-layer network with 4 128-unit wide linear layers for the ensemble methods and a 4-layer network with 4 512-unit wide linear layers for the other methods.
For the MNIST classification problem, we used a 5-layer network with 2 convolution layers and 3 128-unit wide linear layers for our the ensemble methods and a 5-layer network with 2 convolution layers and 3 512-unit wide linear layers for our the other methods.
For the CIFAR classification problem, we used a 7-layer network with 3 convolution layers and 4 256-unit wide linear layers for our the ensemble methods and a 7-layer network with 3 convolution layers and 4 512-unit wide linear layers for our the other methods.

\paragraph{Hyper parameters}
The non-network architecture hyper-parameters we used for GPN were roughly consistent across all experiments, except for the regularization coefficient, $\beta$.
As stated in our experiments section, we performed a small search over 5 values of $\beta$ for each experiment.
The full set of hyper-parameters are shown in Table~\ref{tab:hparams}

\begin{table*}[t]
\begin{center}
\caption{GPN Hyper-parameters used for our experiments.
}
\begin{tabular}{ l c}
\toprule
Hyper-parameter & Value \\
\midrule
\# of sampled embeddings ($k$) & 100 \\
Embedding dimension & 10 \\
Regularization coefficient ($\beta$) & 0.1 \\
Bootstrap network ($\thetab$) & 2-layer linear model \\
Prior variance ($\Sigmab_{\text{prior}}$) & 40 (layer 1), 10 (layer 2) \\
Bootstrap activation & Tanh \\
\bottomrule
\end{tabular}
\label{tab:hparams}
\end{center}
\end{table*}

\input{proofs.tex}

\end{document}

%% file: proofs.tex
\section{Proofs}

\subsection{Proof of Theorem~\ref{thm}}
\label{ap:proof_thm}
\begin{proof}

We will start by showing that $P(\yobsb | \hatYb)$ is normally distributed.

Let $\hatY{i}' = f(\xobs{i}; \thetab)$ be a transformation of the random variable $\thetab$ corresponding to the outputs of the function $f$ parameterized by $\thetab$ evaluated at observation points $\xobsb$.
Note that by definition of $\yobsb$, for each $i \in 1, \ldots, N$:
\begin{align*}
    \yobs{i} = f(\xobs{i}; \thetab) + \epsilon_i = \hatY{i}' + \epsilon_i
\end{align*}
where $\epsilon_i \sim \mathcal{N}(0, \sigma_{\epsilon})$.

By assumption: $[\hatYb, \hatYb']^T$ is normally distributed.
Since $\epsilonb$ is normally distributed and independent of $[\hatYb, \hatYb']^T$, then the joint distribution $[\hatYb, \hatYb', \epsilonb]^T$ is also normally distributed.

Now, by applying a linear transformation, we obtain the joint variable
$
    \begin{bmatrix}
    \hatYb \\ \hatYb' \\ \yobsb
    \end{bmatrix}
    =
    \begin{bmatrix}
    I & 0 & 0 \\
    0 & I & 0 \\
    0 & I & I
    \end{bmatrix}
    \begin{bmatrix}
    \hatYb \\ \hatYb' \\ \epsilonb
    \end{bmatrix}
$, which is also normally distributed (as it is a linear transformation of a Gaussian random variable).

Now, since the conditional and marginal distributions of a Gaussian distribution are also Gaussian, $P(\yobsb, \hatYb' | \hatYb)$ and $P(\yobsb | \hatYb)$ are also Gaussian.

Now, using Bayes' rule, we can show $P(\hatYb | \yobsb)$ is also normally distributed.
\begin{align*}
    P(\hatYb \mid \yobsb)
    &\propto P(\yobsb \mid \hatYb) P(\hatYb) \\
    &= \mathcal{N}( \hatYb | \mub_{\text{like}}, \Sigmab_{\text{like}}) \mathcal{N}(\hatYb | \mub_{\hatYb}, \Sigmab_{\hatYb})
\end{align*}
  for some mean and covariance $\mub_{\text{like}}, \Sigmab_{\text{like}}$.
Since the product of two multivariate Gaussians is a Gaussian, we know that:
\begin{align}
    P(\hatYb \mid \yobsb) = \mathcal{N}(\mub_{\text{post}}, \Sigmab_{\text{post}})
\end{align}
where
\begin{equation}
    \Sigmab_{\text{post}} = \left ( \Sigmab_{\text{like}}^{-1} + \Sigmab_{\hatYb }^{-1} \right)^{-1}, \quad
    \mub_{\text{post}} = \Sigmab_{\text{post}} \Sigmab_{\text{like}}^{-1} \mub_{\text{like}} + \Sigmab_{\text{post}} \Sigmab_{\hatYb }^{-1} \mu_{\hatYb } .
\end{equation}

Now we consider at the distribution $P(\hat{Y}_{\text{MAP}}(\hatYb_{\text{anc}}))$ where $\hatYb_{\text{anc}} \sim \mathcal{N}(\mub_{\text{anc}}, \Sigmab_{\text{anc}})$.
We will show that, when we set
\begin{equation*}
    \mub_{\text{anc}} = \mub_\hatYb, \quad \Sigmab_{\text{anc}} = \Sigmab_\hatYb + \Sigmab_\hatYb \Sigmab_{\text{like}}^{-1} \Sigmab_\hatYb
\end{equation*}
the distribution $P(\hat{Y}_{\text{MAP}}(\hatYb_{\text{anc}}))$ becomes equal to the posterior distribution $\mathcal{N}(\mub_{\text{post}}, \Sigmab_{\text{post}})$.
There are three steps needed to show equality between these distributions:
\begin{enumerate}
    \item Show that $P(\hat{Y}_{\text{MAP}}(\hatYb_{\text{anc}}))$ is normally distributed for some mean and variance, $\mub_{\text{post}}^{\text{RMS}}, \Sigmab_{\text{post}}^{\text{RMS}}$.
    \item Show that $\mub_{\text{post}}^{\text{RMS}} = \mub_{\text{post}}$.
    \item Show that $\Sigmab_{\text{post}}^{\text{RMS}} = \Sigmab_{\text{post}}$.
\end{enumerate}

Using the same reasoning as above, in the limit as $M \rightarrow \infty$,
\begin{align*}
    P(\hat{Y}_{\text{MAP}}(\hatYb_{\text{anc}}))
    &= \argmax_{\hatYb} P(\yobsb | \hatYb) P_{\text{anc}}(\hatYb) \\
    &= \argmax_{\hatYb} \mathcal{N}( \hatYb | \mub_{\text{like}}, \Sigmab_{\text{like}}) \mathcal{N}(\hatYb | \mub_{\text{anc}}, \Sigmab_{\text{anc}})
\end{align*}
Since the max of a Gaussian is the mean, then
\begin{align}
    F_{\text{MAP}}(\hatYb_{\text{anc}}) = A \hatYb_{\text{anc}} + b
\end{align}
where we define:
\begin{align}
    A &= \Sigmab_{\text{post}} \Sigmab_{\hatYb}^{-1} \\
    b &= \Sigmab_{\text{post}} \Sigmab_{\text{like}}^{-1} \mub_{\text{like}}
\end{align}

We will now show that $\mathbb{E}[\hat{Y}_{\text{MAP}}(\hatYb_{\text{anc}})] = \mub_{\text{post}}$.
Because we set $\mub_{\text{anc}} = \mub_{\hatYb}$:
\begin{align*}
    \mathbb{E}[\hat{Y}_{\text{MAP}}(\hatYb_{\text{anc}})]
    &= \mathbb{E}[A \hatYb_{\text{anc}} + b] \\
    &= A \mathbb{E}[\hatYb_{\text{anc}}] + b \\
    &= A \mub_{\hatYb} + b \\
    &= \Sigmab_{\text{post}} \Sigmab_{\hatYb}^{-1} \mub_{\hatYb} + \Sigmab_{\text{post}} \Sigmab_{\text{like}}^{-1} \mub_{\text{like}} \\
    &= \mub_{\text{post}}
\end{align*}

Finally, we will show that $\mathbb{V}\text{ar}[\hat{Y}_{\text{MAP}}(\hatYb_{\text{anc}})] = \Sigmab_{\text{post}}$.
\begin{align*}
    \mathbb{V}\text{ar}[\hat{Y}_{\text{MAP}}(\hatYb_{\text{anc}})]
    &= \mathbb{V}\text{ar}[A \hatYb_{\text{anc}} + b] \\
    &= A \mathbb{V}\text{ar}[\hatYb_{\text{anc}}] A^T \\
    &= (\Sigmab_{\text{post}} \Sigmab_{\hatYb}^{-1}) (\Sigmab_{\hatYb} + \Sigmab_{\hatYb} \Sigmab_{\text{like}}^{-1} \Sigmab_{\hatYb}) (\Sigmab_{\text{post}} \Sigmab_{\hatYb}^{-1})^T \\
    &= (\Sigmab_{\text{post}}  + \Sigmab_{\text{post}}  \Sigmab_{\text{like}}^{-1} \Sigmab_{\hatYb}) (\Sigmab_{\hatYb}^{-1} \Sigmab_{\text{post}} ) \\
    &= \Sigmab_{\text{post}} \Sigmab_{\hatYb}^{-1} \Sigmab_{\text{post}}  + \Sigmab_{\text{post}}  \Sigmab_{\text{like}}^{-1} \Sigmab_{\text{post}}\\
    &= \Sigmab_{\text{post}} (\Sigmab_{\hatYb}^{-1} +  \Sigmab_{\text{like}}^{-1} ) \Sigmab_{\text{post}}\\
    &= \Sigmab_{\text{post}}\\
\end{align*}

So, $P(\hat{Y}_{\text{MAP}}(\hatYb_{\text{anc}})) \sim \mathcal{N}(\mub_{\text{post}}^{\text{RMS}}, \Sigmab_{\text{post}}^{\text{RMS}})$ where $\mub_{\text{post}}^{\text{RMS}} = \mub_{\text{post}}$ and $\Sigmab_{\text{post}}^{\text{RMS}} = \Sigmab_{\text{post}}$.

Thus, in the limit as $M \rightarrow \infty$, $P(\hat{Y}_{\text{MAP}}(\hatYb_{\text{anc}})) = P(\hatYb | \yobsb)$.

\end{proof}

\subsection{Derivation of loss function}
\label{ap:proof_loss}

We start with
\begin{align*}
    \phib
    &= \argmax_{\phib} \mathbb{E}_{\thetab_{\text{anc}} \sim P_{\text{prior}}(\thetab_{\text{anc}})} \log P_{\text{like}}(\yobsb \mid g(\xsampleb, \thetab_{\text{anc}}; \phib)) + \log P_{\text{anc}} (g(\xsampleb, \thetab_{\text{anc}}; \phib))
\end{align*}

If we choose $\mub_{\text{anc}} = \mub_{\hatYb}$ and $\Sigmab_{\text{anc}} = \Sigmab_{\hatYb}$, we can simplify the logarithm of the likelihood and anchor distributions as follows:
\begin{align*}
    \log P_{\text{like}}(\yobsb \mid g(\xsampleb, \thetab_{\text{anc}}; \phib))
    &= \sum_i \log \mathcal{N}(\yobs{i} | g(\xsample{i}, \thetab_{\text{anc}}; \phib), \sigma_{\epsilon}) \\
    &= -\frac{1}{2 \sigma_{\epsilon}^2} \sum_i  \| \yobs{i} - g(\xsample{i}, \thetab_{\text{anc}}; \phib) \|_2^2
\end{align*}
\begin{align*}
    \log P_{\text{anc}} (g(\xsampleb, \thetab_{\text{anc}}; \phib))
    &= \log \mathcal{N}(g(\xsampleb, \thetab_{\text{anc}}; \phib) | \mub_{\hatYb}, \Sigmab_{\hatYb}) \\
    &= - \frac{1}{2} \deltab^T \Sigmab_{\hatYb}^{-1} \deltab
\end{align*}
where $\delta_j = g(\xsample{j}, \thetab_{\text{anc}}; \phib) - f(\xsample{j}; \thetab_{\text{anc}})$.

Putting these together, we get:
\begin{align*}
    \phib
    &= \argmax_{\phib} \mathbb{E}_{\thetab_{\text{anc}} \sim P_{\text{prior}}(\thetab_{\text{anc}})} -\frac{1}{2 \sigma_{\epsilon}^2} \sum_i \| \yobs{i} - g(\xsample{i}, \thetab_{\text{anc}}; \phib) \|_2^2 - \frac{1}{2} \deltab^T \Sigmab_{\hatYb}^{-1} \deltab \\
    &= \argmin_{\phib} \mathbb{E}_{\thetab_{\text{anc}} \sim P_{\text{prior}}(\thetab_{\text{anc}})} \frac{1}{N} \sum_i  \| \yobs{i} - g(\xsample{i}, \thetab_{\text{anc}}; \phib) \|_2^2 + \frac{1}{N} \sigma_{\epsilon}^2 \deltab^T \Sigmab_{\hatYb}^{-1} \deltab
\end{align*}